\documentclass[lettersize,journal]{IEEEtran}
\IEEEoverridecommandlockouts

\usepackage{graphicx}
\usepackage{amsmath,amssymb}
\usepackage{bm}
\usepackage{cite}
\usepackage{dsfont}
\usepackage{placeins}
\usepackage{subcaption}
\usepackage{xurl}

\begin{document}

	\title{
        Zero-Shot Policy Transfer in Reinforcement Learning using Buckingham's Pi Theorem
	}

    \author{Francisco Pascoa$^1$, Ian Lalonde$^1$ and Alexandre Girard$^1$ 

        \thanks{$^1$ F. Pascoa, I. Lalonde and A. Girard are with the Department of Mechanical Engineering, Université de Sherbrooke, Qc, Canada.}
    }

    \markboth{This work has been submitted to the IEEE for possible publication. Copyright may be transferred without notice.}{}

	\maketitle{}

	\begin{abstract}
    Reinforcement learning (RL) policies often fail to generalize to new robots, tasks, or environments with different physical parameters, a challenge that limits their real-world applicability.
    This paper presents a simple, zero-shot transfer method based on Buckingham's Pi Theorem to address this limitation.
    The method adapts a pre-trained policy to new system contexts by scaling its inputs (observations) and outputs (actions) through a dimensionless space, requiring no retraining.
    The approach is evaluated against a naive transfer baseline across three environments of increasing complexity: a simulated pendulum, a physical pendulum for sim-to-real validation, and the high-dimensional HalfCheetah.
    Results demonstrate that the scaled transfer exhibits no loss of performance on dynamically similar contexts.
    Furthermore, on non-similar contexts, the scaled policy consistently outperforms the naive transfer, significantly expanding the volume of contexts where the original policy remains effective.
    These findings demonstrate that dimensional analysis provides a powerful and practical tool to enhance the robustness and generalization of RL policies.
	\end{abstract}

    \section{Introduction}

Reinforcement learning (RL) in robotics has proven itself useful to synthesize controllers for complex, non-linear systems.
However, two fundamental challenges limit its deployment in real-world applications.
First, RL is sample-inefficient: algorithms can require millions of environment interactions to converge to a high-performing policy.
On physical hardware, such extensive trial-and-error is impractical due to time constraints, mechanical wear, and safety concerns.
Second, RL policies lack robustness.
They are typically trained on a single environment—often in simulation—and are brittle when conditions deviate from the original context.

To address the issue of data efficiency, previous work first developed better off-policy algorithms such as SAC \cite{haarnoja_soft_2018}, TD3 \cite{fujimoto_addressing_2018}, and TQC \cite{kuznetsov_controlling_2020}.
However, the improvements of these algorithms are at the cost of complexity.
Most of them introduce additional neural networks (NN) with elaborate update schedules.
Going against this trend, the most recent development on that front, CrossQ \cite{bhatt_crossq_2024}, improves upon SAC by simplifying its architecture.
It results in better sample and computational efficiency as well as better performance.
This highlights that architectural simplification can improve both sample efficiency and computational cost.

Another strategy, called transfer learning (TL), is to reduce the amount of data necessary by adapting a pre-trained policy trained in a different context (robot, task or environnement).
A survey of TL in RL \cite{zhu_transfer_2023} summarizes the approaches in five categories: reward shaping, learning from demonstration, policy transfer, inter-task mapping, and representation transfer.
Despite extensive research, there are still open problems in the field of transfer learning: 1) the search for fundamental representations of knowledge, and 2) a systematic way to quantify the similarity between two situations to predict the quality of a transfer \cite{jaquier_transfer_2025}.

Alternatively, physics-informed learning (PIL) can also augment data efficiency and robustness by exploiting prior physical knowledge about the task to better model what needs to be learned.
Examples of PIL range from a new NN architecture to learn the Lagrangian of a system \cite{cranmer_lagrangian_2020} to leveraging symmetries in the task \cite{welde_leveraging_2025}.
Insight from PIL has the potential to allow TL in a simple manner.

Dimensional analysis and Buckingham's Pi theorem (BPT) \cite{buckingham_physically_1914} lies at the intersection of PIL and TL.
While widely used in fields like fluid mechanics to generalize experimental results, dimensional analysis is a largely untapped and overlooked tool in robotics and control theory.
Mixing BPT with neural networks (NN) was first studied by Gunaratnam et al.\cite{gunaratnam_improving_2003},
where the authors aimed to reduce the size of NNs by further combining dimensionless quantities in supervised learning situations.
Though they didn't look at TL, their results showed that BPT improved performance with smaller NNs.
A similar approach was later explored by Oppenheimer et al. \cite{oppenheimer_multi-scale_2022}.
In supervised learning, the authors trained a NN in the dimensionless space on one system and showed that it could effectively work on a previously unseen system.
They emphasize that for the transfer to work, the dimensionless input domain of the new system must lie within the training input domain.
This indicates that to hope for a successful policy transfer, the observations produced by the new system must fit within the range of observations seen by the policy during training.

Makarun et al.~\cite{makarun_testing_2021} explored the transfer of policies with BPT.
They used BPT to match a physical small-scale vehicle to a full-scale simulation.
They then tuned a controller on the small scale and deployed it on the full scale.
The transfer was mostly successful but required further tuning on the full scale.
This discrepancy can be attributed to some parameters of the physical small-scale vehicle that could not be matched exactly.
Although the controller wasn't a NN, it agrees with the conclusion of Oppenheimer et al. \cite{oppenheimer_multi-scale_2022}---the dimensionless input domain of the small-scale system was not the same as that of the full-scale system.

Lastly, our previous work formalize a framework for dimensionless policy and brings a theoretical result concluding that a single optimal, dimensionless policy can be scaled for reuse across any dimensionally similar system without performance loss \cite{girard_dimensionless_2024}.
Charvet et al. \cite{charvet_improving_2025} extend this idea by training the policy on dimensionless observations. They observe that the resulting policy is more robust to context perturbations than a completely dimensional policy.

While prior work has established a theoretical foundation for dimensionless policy transfer and has shown the benefits of training policies directly on dimensionless data, a practical gap remains. The existing theoretical guarantees apply to cases of perfect dimensional similarity, and empirical studies have been limited to simpler systems or non-NN controllers. It remains an open question whether these principles can be applied to transfer existing, pre-trained
dimensional policies to complex, high-dimensional systems or across the sim-to-real gap, especially when similarity is not exact.

This paper addresses these open questions through several novel contributions. It provides a practical investigation into the zero-shot transfer of existing, pre-trained dimensional policies from state-of-the-art RL algorithms. The method's viability is then assessed under challenging conditions: its scalability is tested on the high-dimensional \textit{HalfCheetah} system, and its practicality for real-world applications is validated through a sim-to-real transfer on a physical pendulum. Finally, the analysis extends beyond ideal conditions to evaluate performance on quasi-similar contexts. This analysis demonstrates the method's effectiveness when dimensional similarity is not exact, showing that dimensionless scaling consistently provides a larger region of viable transfer than a naive approach and allowing for a quantitative assessment of the method's limits as contexts similarity diverge.

    \FloatBarrier
    \section{Theory}\label{sec:theory}

\subsection{Buckingham's Pi Theorem}\label{sec:BPT}

Quantities are numerical values coupled with a dimension that describes a system.
This paper focuses on dynamical systems; therefore, the dimension space is mass-length-time (M L T).
The dimension of a quantity $q$ is usually denoted with square brackets \cite{sonin_physical_2004},
\begin{equation}
	[q] = M^a L^b T^c \equiv (a, b, c)
\end{equation}
where $a, b, c \in \mathds{R}$.

Let there be a dynamical system that follows:
\begin{equation}
	f(q_1, q_2, ..., q_n) = 0 \,.
	\label{eq sys}
\end{equation}
Buckingham's Pi theorem states that \eqref{eq sys} can be rewritten as its dimensionless form:
\begin{equation}
	F(\tilde{q}_1, \tilde{q}_2, ..., \tilde{q}_{n-3}) = 0
	\label{eq adim sys}
\end{equation}
where:
\begin{equation}
	\tilde{q}_j = \phi_\beta(q_i) = q_i \prod_{q_b \in \beta} q_b^{m_b}
	\label{eq pi prod}
\end{equation}
with $i \neq b$ and where $\beta \subseteq \{q_1, q_2, ..., q_n\}$ is the basis such that the dimension of the quantities spans the dimension space. Therefore, $\phi_\beta$ is the transformation of a quantity to its dimensionless counterpart using basis $\beta$, while $\phi_\beta^{-1}$ transforms a dimensionless quantity to its dimensional quantity from a given context.
The powers $m_b$ are given by:
\begin{equation}
	\bm[ m_{b_1} ~ m_{b_2} ~ m_{b_3} \bm]^T = -\bm[\, [q_{b_1}] ~ [q_{b_2}] ~ [q_{b_3}] \,\bm]^{-1} [q_i] \,.
\end{equation}
The tilde is used to denote a dimensionless quantity, reserving $\pi$ as the symbol for policies.

The key property from equation \eqref{eq adim sys} is that two physical systems yielding $\tilde{q}_j$, that is, of the same values, behave the same way independently of scale.
This property provides the foundation of the proposed approach.

\subsection{Context Similarity}

Let $\pi$ be a policy such that
\begin{equation}
    \pi(\bm{x}) \equiv \pi(\bm{x}, \mathcal{C})
\end{equation}
where $\bm{x}$ is the current state of the system and $\mathcal{C}$ is the \emph{context}.
The context of a system is the set of quantities that are implicit parameters embedded in the policy.
The quantities include the system's parameters, but also the parameters of the cost function that the policy optimizes and the parameters of any other ``structure'' that is used to produce the policy---such as the time step of a simulator.

Two contexts can be compared using their dimensionless distance.
Let $\mathcal{C}_1$ and $\mathcal{C}_2$ be two different contexts and
$\beta_i$ be the basis with values that define the context $\mathcal{C}_i$
Then the distance between the two contexts is given by:
\begin{equation}
    d_\beta(\mathcal{C}_1, \mathcal{C}_2) = ||\phi_{\beta_2}(\mathcal{C}_2) - \phi_{\beta_1}(\mathcal{C}_1)||
    \label{eq dimensionless distance}
\end{equation}
where $\phi_{\beta_i}(\mathcal{C}_i)$ gives the dimensionless context by applying the transformation to each of its elements and $||\cdot||$ is the Euclidean distance.
Using equation \eqref{eq dimensionless distance}, two contexts are said to be \emph{similar} if:
\begin{equation}
    d_\beta(\mathcal{C}_1, \mathcal{C}_2) = 0 \,.
\end{equation}

\subsection{Policy Scaling}\label{sec:scaling}

The transfer of a policy $\pi$ from $\mathcal{C}_1$ to $\mathcal{C}_2$ via the basis $\beta$ is called \emph{scaling} and corresponds to:
\begin{equation}
    ^{\mathcal{C}_1}\pi_\beta^{\mathcal{C}_2}(\bm{x}) = (\phi_{\beta_2}^{-1} \circ \phi_{\beta_1}) \circ \pi \circ (\phi_{\beta_1}^{-1} \circ \phi_{\beta_2}) (\bm{x}) \,.
\end{equation}
For brevity, a policy scaled to a new context using BPT will be referred to as a \emph{scaled policy}.

According to previous work \cite{girard_dimensionless_2024, charvet_improving_2025} and the key property from Section~\ref{sec:BPT}, if $\mathcal{C}_1$ and $\mathcal{C}_2$ are similar and $\pi(\bm{x}, \mathcal{C}_1)$ is optimal for $\mathcal{C}_1$, then $^{\mathcal{C}_1}\pi_\beta^{\mathcal{C}_2}(\bm{x})$ remains optimal for $\mathcal{C}_2$.

    \FloatBarrier
    \section{Objective and methodology}

The objective of this study is to determine the circumstances in which BPT enables a policy to be transferred.
To achieve this, three environments of progressive complexity are considered: a simulated simple pendulum, a real physical simple pendulum and Gymnasium's \cite{towers_gymnasium_2023} \textit{HalfCheetah}.

\subsection{Overarching methodology}

The core methodology is the same across the environments.
First, a policy is obtained for the \emph{original context} of the environment---the \emph{original policy}.
Second, a test region around the original context is defined by systematically varying its physical parameters.
Finally, the performance of two transfers of the original policy is evaluated:
\begin{itemize}
    \item naive transfer -- the original policy is used as is on different contexts;
    \item scaled transfer -- the original policy is scaled as described in Section~\ref{sec:theory}.
\end{itemize}
The performance is evaluated through the reward function of the environments.

\subsection{Simulated Inverted Pendulum}\label{sec:sim pendulum}

The simple pendulum is studied for its simple dynamics and small number of parameters.
Table \ref{tab contexte pendule} presents the context variables of the pendulum with the transformation function $\phi_\beta$ for each quantity $(q)$, and Table \ref{tab obs act pendule} presents its observation and action spaces.

\begin{table}[htb]
	\centering
	\setlength{\tabcolsep}{5pt}
    \renewcommand{\arraystretch}{1.8}
	\begin{tabular}{l|c|c|c}
		Quantity    & Description              & Dimension               & $\phi_\beta(q)$               \\
		\hline\hline
		$m$           & Mass                     & M                       & 1                             \\
		$g$           & Gravitational acceleration & L T$^{-2}$            & 1                             \\
		$l$           & Pendulum length          & L                       & 1                             \\
		$\tau_{\max}$ & Maximum torque           & M L$^2$ T$^{-2}$        & $\tau_{\max} \frac{1}{mgl}$   \\
		$t_f$         & Episode duration         & T                       & $t_f \sqrt\frac{g}{l}$        \\
		$w_\theta$    & Cost weight on position  & T$^{-1}$                & $w_\theta \sqrt\frac{l}{g}$   \\
		$w_\tau$      & Cost weight on torque    & M$^{-2}$ L$^{-4}$ T$^3$ & $w_\tau m^2 \sqrt{g^3 l^5}$   \\
	\end{tabular}
	\caption{Simulated pendulum context with transformations. $\beta = \{m, l, g\}$.}
	\label{tab contexte pendule}
\end{table}

\begin{table}[htb]
    \centering
    \begin{tabular}{c|l|c|c}
        Quantity        & Description              & Dimension                 & $\phi_\beta(q)$                \\
        \hline\hline
        \multicolumn{4}{c}{Observations} \\
        \hline
        $\theta$      & Angular position         & 1                         & $\theta$                        \\
        $\dot\theta$  & Angular velocity         & T$^{-1}$                  & $\dot\theta \sqrt{\tfrac{l}{g}}$     \\
        \hline
        \multicolumn{4}{c}{Actions} \\
        \hline
        $\tau$        & Applied torque           & M L$^2$ T$^{-2}$          & $\dfrac{\tau}{m g l}$          \\
    \end{tabular}
    \caption{Observations and actions of the simple pendulum. $\beta = \{m, l, g\}$.}
    \label{tab obs act pendule}
\end{table}

The original context $\mathcal{C}_0$ is defined by the values:
\begin{equation}
\begin{aligned}
\mathcal{C}_0 = \{\,
& m_0 = 1 \ \text{kg}, \quad g_0 = 10 \ \text{m s}^{-2}, \quad l_0 = 2 \ \text{m}, \\
& \tau_{\max,0} = 8 \ \text{N m}, \quad t_{f,0} = 10 \ \text{s}, \\
& w_{\theta,0} = 1 \ \text{s}^{-1}, \quad w_{\tau,0} = 0.01 \ \text{N}^{-2}\text{s}
\,\}.
\end{aligned}
\end{equation}
The test contexts $\mathcal{C}_{\text{t}}$ are given by:
\begin{equation}
    \mathcal{C}_{\text{t}} = \phi_{\beta_t}^{-1} \circ \phi_{\beta_0} (\mathcal{C}_{\text{t}})
    \label{eq scale context}
\end{equation}
with $m_t \in [m_0 / 10,~ 10\,m_0]$, $l_t \in [l_0 / 10,~ 10\,l_0]$, $g_t = g_0$ and by varying $\tau_{\max_t} \in [\tau_{\max_0} / 10,~ 10\,\tau_{\max_0}]$ without regard to similarity.

The original policy is a lookup table computed by dynamic programming.
Table \ref{tab obs act pendule} presents its observation and action spaces.
The basis used for the transformation is $\beta = \{m, l, g\}$.
The scaled policy is given by:
\begin{equation}
    ^{\mathcal{C}_0}\pi_\beta^{\mathcal{C}_t}(\theta_t, \dot\theta_t) = \frac{m_t l_t g_t}{m_0 l_0 g_0} \pi\left(\theta_t, \dot\theta_t \sqrt\frac{l_t g_0}{g_t l_0}\right) \,.
\end{equation}

The reward function of the environment is:
\begin{equation}
    \text{reward}(\theta, \tau) = -\int_{0}^{t_f} ( w_\theta \theta(t)^2 + w_\tau \tau(t)^2 ) dt \,.
    \label{eq reward pendule}
\end{equation}
The reward function is chosen---without loss of generality---to be dimensionless, which explains the dimensions of $w_\theta$ and $w_\tau$ in Table \ref{tab contexte pendule}.

\subsection{Real Inverted Pendulum}

The real pendulum is assembled using an AK60-6 Cubemars motor and a 610~mm long carbon tube with a rectangular cross-section of 51~mm by 25~mm, allowing weights to be attached at various positions along the tube.
The real pendulum serves to validate the theory in a sim-to-real situation. The context, observation, and action variables are the same as the simulated pendulum and are thus defined in table~\ref{tab contexte pendule} and table~\ref{tab obs act pendule}.

The original context $\mathcal{C}_0$ is defined by the values:
\begin{equation}
\begin{aligned}
\mathcal{C}_0 = \{\,
& m_0 = 1.07 \ \text{kg}, &\quad g_0 = 9.81 \ \text{m s}^{-2}, \\
&\quad l_0 = 0.37 \ \text{m}, & \tau_{\max,0} = 1.6 \ \text{N m}, &\quad t_{f,0} = 6 \ \text{s}, \\
& w_{\theta,0} = 1 \ \text{s}^{-1}, &\quad w_{\tau,0} = 0.01 \ \text{N}^{-2}\text{s}
\,\}.
\end{aligned}
\end{equation}

The test contexts $\mathcal{C}_t$ are generated in the same fashion as the simulated pendulum.
The mass $m \in \{0.74, 1.07, 1.73\}$ is adjusted with metal plates, the length $l \in \{0.31, 0.37, 0.44\}$ by sliding the mass along the arm and $\tau_{\max} \in \{0.93, 1.6, 3.1\}$ is limited in software.
The contexts with $(m=0.74, l=0.31, \tau_{\max}=0.93)$ and $(m=1.73, l=0.44, \tau_{\max}=3.1)$ are chosen to be similar to $\mathcal{C}_0$.

The original policy is a Stable-Baselines3 \cite{raffin_stable-baselines3_2021} SAC model trained in simulation with the default hyperparameters, a training time of 30000 steps, and an episode length of 400 steps.

All the other details are the same as the simulated pendulum described in Section~\ref{sec:sim pendulum}.

\subsection{HalfCheetah}

The HalfCheetah is used to validate that policy scaling can be generalized to high-dimensional systems with complex dynamics
Table \ref{tab:contexte_half_cheetah} presents the original context $\mathcal{C}_0$.
The basis used for the transformation is $\beta = \{m, l_0, g\}$.

\begin{table}[htb]
    \centering
    \renewcommand{\arraystretch}{1.8}
    \begin{tabular}{c|l|c|c}
        Quantity      & Description                          & Dimension                 & $\phi_\beta(q)$                                \\
        \hline\hline
        $m$           & Total mass                             & M                         & 1                                 \\
        $g$           & Gravitational acceleration             & L T$^{-2}$                & 1                                 \\
        $\tau_{\max}$ & Maximum torque                          & M L$^2$ T$^{-2}$          & $\frac{\tau_{\max}}{m l_0 g}$     \\
        $d$           & Link diameter                           & L                         & $\frac{d}{l_0}$                   \\
        $L$           & Body length                              & L                         & $\frac{L}{l_0}$                   \\
        $L_h$         & Head length                               & L                         & $\frac{L_h}{l_0}$                 \\
        $l_0$         & Reference leg length                     & L                         & 1                                 \\
        $l_i$         & i-th leg segment length                  & L                         & $\frac{l_i}{l_0}$                 \\
        $k_i$         & i-th joint stiffness                     & M L$^2$ T$^{-2}$          & $\frac{k_i}{m l_0 g}$             \\
        $b_i$         & i-th joint damping                       & M L$^2$ T$^{-1}$          & $\frac{b_i}{m \sqrt{l_0^3 g}}$    \\
        $I_m$         & Motor inertia                             & M L$^2$                   & $\frac{I_m}{m l_0^2}$             \\
        $w_f$         & Forward reward weight                     & L$^{-1}$ T                & $w_f \sqrt\frac{g}{l_0^3}$        \\
        $w_c$         & Control cost weight                       & M$^{-2}$ L$^{-4}$ T$^4$   & $w_c m^2 l_0^2 g^2$               \\
    \end{tabular}
    \caption{HalfCheetah context with descriptions and dimensionless transformations. $\beta = \{m, l_0, g\}$; $i \in \{1, 2, 3, 4, 5\}$ denotes the joints.}
    \label{tab:contexte_half_cheetah}
\end{table}

The test contexts $\mathcal{C}_t$ are generated following equation \eqref{eq scale context} with $m_t \in [m_0 / 2, 2 m_0]$, $l_{0,t} \in [l_{0,0} / 2, 2 l_{0,0}]$, $g_t = g_0$ and by varying body length $L_t \in [L_0 / 2, 2 L_0]$ without regard to similarity.

\begin{table}[htb]
    \centering
    \renewcommand{\arraystretch}{1.8}
    \begin{tabular}{c|l|c|c}
        Quantity        & Description                          & Dimension                 & $\phi_\beta(q)$                                \\
        \hline\hline
        \multicolumn{4}{c}{Observations} \\
        \hline
        $z$             & Vertical position of the body         & L                         & $\frac{z}{l_0}$                   \\
        $\psi$          & Body pitch angle                       & 1                         & $\psi$                            \\
        $\theta_i$      & i-th joint angle                        & 1                         & $\theta_i$                        \\
        $\dot{x}$       & Forward velocity of the body            & L T$^{-1}$                & $\frac{\dot{x}}{\sqrt{l_0 g}}$     \\
        $\dot{z}$       & Vertical velocity of the body           & L T$^{-1}$                & $\frac{\dot{z}}{\sqrt{l_0 g}}$     \\
        $\dot{\psi}$    & Angular velocity of body pitch          & T$^{-1}$                  & $\dot{\psi} \sqrt{\frac{l_0}{g}}$     \\
        $\dot{\theta}_i$& Angular velocity of i-th joint          & T$^{-1}$                  & $\dot{\theta}_i \sqrt{\frac{l_0}{g}}$ \\
        \hline
        \multicolumn{4}{c}{Actions} \\
        \hline
        $\tau_i$        & Torque applied at i-th joint           & M L$^2$ T$^{-2}$          & $\frac{\tau_i}{m g l_0}$          \\
    \end{tabular}
    \caption{HalfCheetah observations and actions with descriptions and dimensionless transformations. $i \in \{0,1,2,3,4,5\}$ denotes individual joints.}
    \label{tab:obs_act_half_cheetah}
\end{table}

The original policy is a state-of-the-art pre-trained TQC model \cite{kallinteris_andreas_farama-minarihalfcheetah-v5-tqc-expert_2025}.
Table~\ref{tab:obs_act_half_cheetah} presents its observation and action spaces.
The scaled policy is given by:
\begin{equation}
    ^{\mathcal{C}_0}\pi_\beta^{\mathcal{C}_t}(\bm{s}_t) = \frac{m_t l_{0,t} g_t}{m_0 l_{0,0} g_0} \pi(\bm{s}_0(\bm{s}_t)) \,.
\end{equation}
with
\begin{equation}
    \bm{s}_t = (z_t, \psi_t, \theta_{i,t}, \dot{x}_t, \dot{z}_t, \dot\psi_t, \dot\theta_{i,t})
\end{equation}
and

\begin{equation}
\begin{aligned}
\bm{s}_0(\bm{s}_t) = \bigg(&
z_t \frac{l_{0,0}}{l_{0,t}}, \
\psi_t, \
\theta_{i,t},
 \dot{x}_t \sqrt{\frac{l_{0,0} g_0}{l_{0,t} g_t}}, \\ \
&\dot{z}_t \sqrt{\frac{l_{0,0} g_0}{l_{0,t} g_t}},
 \dot{\psi}_t \sqrt{\frac{l_{0,t} g_0}{g_t l_{0,0}}}, \
\dot{\theta}_{i,t} \sqrt{\frac{l_{0,t} g_0}{g_t l_{0,0}}}
\bigg)
\end{aligned}
\end{equation}

The reward function is defined in the environment:
\begin{equation}
    \text{reward}(\dot{x}, \tau) = \int_{0}^{t_f} ( w_f \dot{x}(t) - w_c \tau(t)^2 ) dt \,.
    \label{eq reward pendule}
\end{equation}
As with the pendulum, the coefficients $w_f$ and $w_c$ are chosen such that the overall reward is dimensionless.

    \FloatBarrier
    \section{Results}

Before presenting the results, one important clarification is needed.
BPT ensures that for similar contexts, the total reward is directly comparable.
However, this statement does not hold for non-similar contexts, meaning that the performance in one context cannot be qualified as better or worse than in another non-similar context.
Nonetheless, within the same context, the total reward of the scaled policy and the naive policy remains perfectly comparable.

The experiments conducted on the three environments show that the scaled transfer policy performs better than the naive transfer policy.

\begin{figure}[htb]
	\centering
	\includegraphics[width=1.0\linewidth]{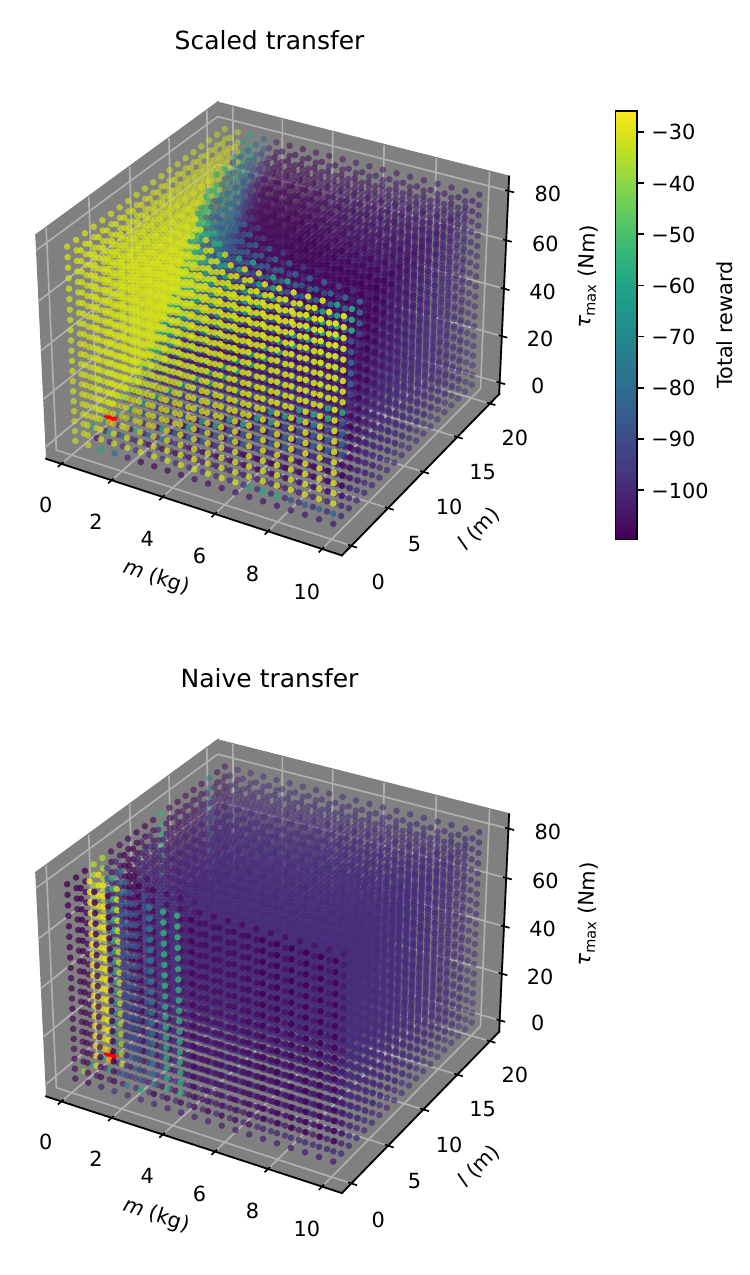}
	\caption{
        The scaled policy yields a higher total reward on more contexts than the naive policy on the simulated pendulums.
        Each dot represents the total reward of the policy in the given context for scaled and naive transfer.
        The red star marks the original context.
        The basis used is $\beta = \{m, l, g\}$.
    }
	\label{fig pendule volume}
\end{figure}

Figure~\ref{fig pendule volume} compares the total rewards of the scaled policy and naive policy on simulated pendulums.
The results show that the scaled policy achieves a higher total reward on a wider range of contexts than the naive policy.
As discussed in section~\ref{sec:scaling}, scaling the policy ensures that contexts similar to the original achieve the same reward as the original context.
For both policies, contexts with more available torque also achieve the same reward as the original, since the policy does not exploit the excess torque.

\begin{figure}[htb]
    \centering
    \includegraphics[width=1.0\linewidth]{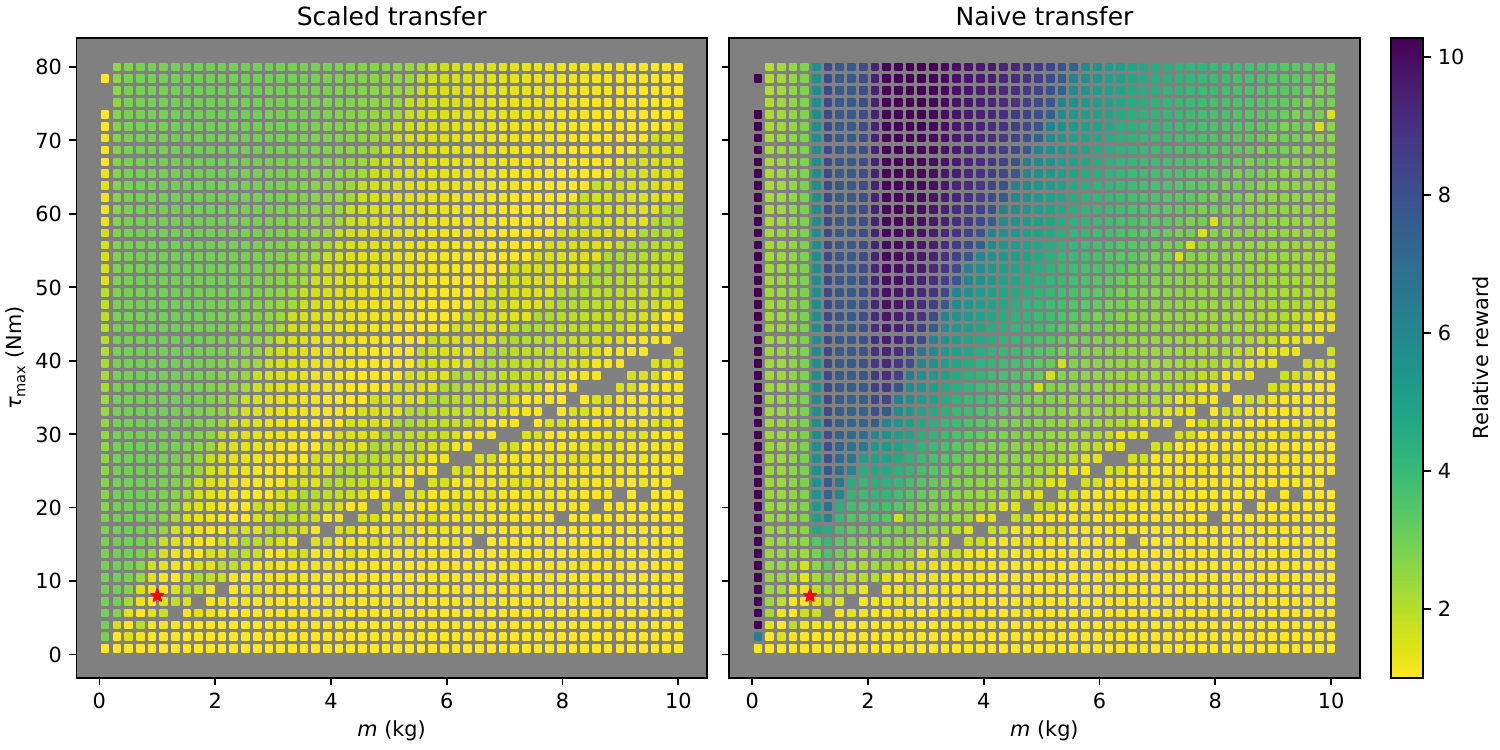}
    \caption{
        The scaled policy remains optimal on the diagonal where similar contexts to the original lie.
        Each dot represents the total reward of the policy in the given context for scaled and naive transfer.
        The red star marks the original context.
        The basis used is $\beta = \{m, l, g\}$.
    }
    \label{fig pendule relatif}
\end{figure}

The contexts from Figure~\ref{fig pendule relatif} correspond to a vertical slice of Figure~\ref{fig pendule volume} at the original length.
The relative reward is defined as the ratio between the total reward of the tested policy and the reward of the optimal policy for the context.
The results show that for contexts similar to the original (along the diagonal), the scaled policy is optimal.
For non-similar contexts, the scaled policy is no longer optimal; however, it outperforms the naive policy.

\begin{figure}[htb]
    \centering
    \includegraphics[width=1.0\linewidth]{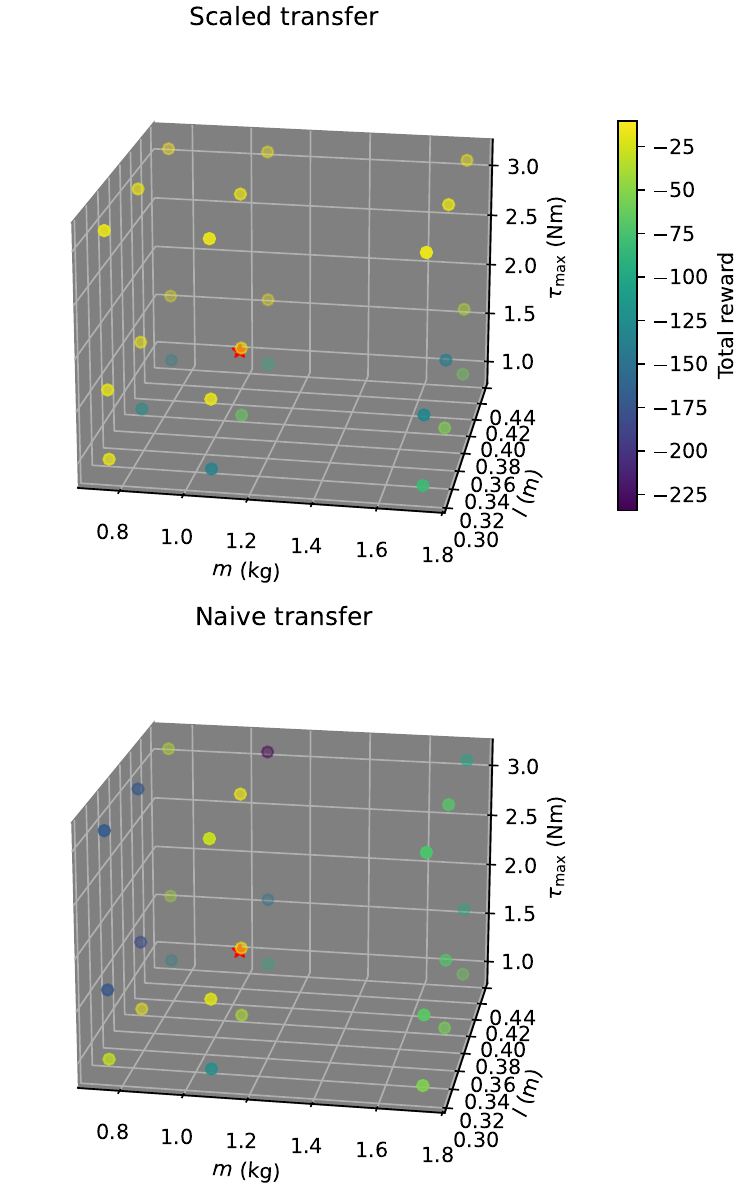}
    \caption{
        The scaled policy yields a higher total reward on more contexts than the naive policy on a real pendulum.
        Each dot represents the total reward of the policy in the given context for scaled and naive transfer.
        The red star marks the original context.
        Contexts on the main diagonal (front lower left to back upper right) a similar to the original.
        The basis used is $\beta = \{m, l, g\}$.
    }
    \label{fig result pendule irl}
\end{figure}

Figure \ref{fig result pendule irl} is analogous to Figure \ref{fig pendule volume} for the real pendulum transfer.
It shows that a scaled policy also yields better performances in sim-to-real situations.

\begin{figure}[htb]
    \centering
    \includegraphics[width=1.0\linewidth]{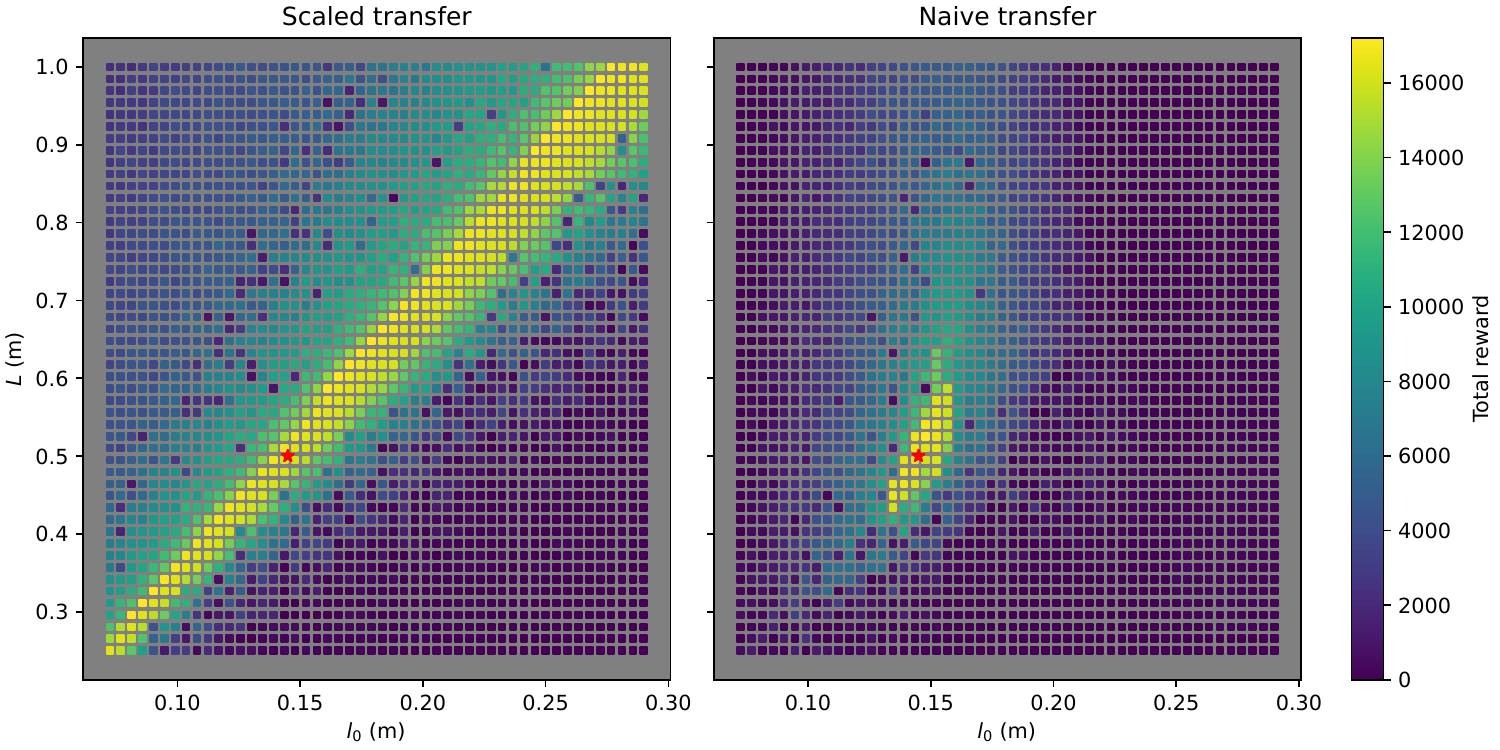}
    \caption{
        The scaled policy yields a higher total reward on more contexts than the naive transfer on the HalfCheetah.
        Each dot represents the total reward of the policy in the given context for scaled and naive transfer.
        The red star marks the original context.
        Contexts on the diagonal are perfectly similar to the original.
        Only the body length $L$ is not similar. $l_0$ is the back thigh length.
        The basis used is $\beta = \{m, l_0, g\}$.
    }
    \label{fig half-cheetah resultat}
\end{figure}

Figure \ref{fig half-cheetah resultat} presents a comparison of the total reward scored by the scaled policy and the naive policy on the HalfCheetah.
The scaled policy outperforms the naive policy across a broader range of contexts.
It also shows that on contexts similar to the original, the scaled policy achieves the same performance as in the original environment.

\section{Discussion}

Results confirm that zero-shot transfer using scaled policies works perfectly on contexts similar to the original.
Experimental tests with a real pendulum confirm that this result generalizes across the sim-to-real barrier.
This success is because on dynamically similar contexts, the BPT scaling ensures that from the policy's perspective, the system's observations and actions are the same as the original context.

For the non-similar contexts, a scaled policy is no longer guaranteed to be optimal.
Some changes in the context are transparent to the basis. In the case of the HalfCheetah, variations in the body length are not captured by the chosen basis.
Therefore, in those cases, the scaled policy is effectively transferred naively to this non-similar context from the closest similar context.
This scaling to a more similar context results in a generally better performance than a completely naive policy transfer.
This means, when training an RL, that a scaled policy can serve as a better initial guess when retraining for a new system and save on retraining time.

The performance of the naive policy across different contexts indicates in which direction of the context space the original policy is naturally robust.
Accordingly, the scaled policy is also robust along these directions.
This can be observed in both Figure \ref{fig pendule volume} and Figure \ref{fig half-cheetah resultat}.
For the pendulum, having more torque available does not degrade the performance of the policy.
For the HalfCheetah, a longer body is less prone to flipping over than a short body and thus more stable.
Knowing the direction in which the policy is naturally stable can help predict the robustness of the scaled policy, since the naive policy's robustness region is scaled to other contexts.
This is observed in Figure \ref{fig half-cheetah resultat} as the small area of high-performing contexts under the naive transfer is expanded into a larger area along the diagonal of similar contexts with the scaled transfer.

    \FloatBarrier
    \section{Conclusion}

Given a policy trained on an original context, Buckingham's Pi theorem enables zero-shot transfer to dynamically similar contexts.
On non-similar contexts, a scaled policy is not guaranteed to be optimal, but generally performs better than a naive transfer.
When further training is necessary, a scaled policy provides a strong initial guess, reducing retraining time and resources needed to optimize the policy.
In any case, the robustness of the original policy dictates the robustness of the scaled policy and is scaled accordingly.
Experimental validations confirm that these theoretical observations extend beyond simulation and hold in real-world scenarios.

Future work could replicate the experimental study on a more complex system.
Furthermore, a sim-to-real transfer rarely yields a perfectly similar context to the original, and a real context can also vary dynamically during the robot's operation.
A study of the effectiveness of a context estimator to improve the robustness of a policy in the real world would be valuable.

    ~\\
    \textbf{Data availability statement:} The code used to generate the results in this paper is available at: \url{https://github.com/SherbyRobotics/zero_shot_buckingham_pi_transfer}



\FloatBarrier
\bibliographystyle{IEEEtran}
\bibliography{references}

\end{document}